\documentclass[letterpaper, 10 pt, conference]{ieeeconf}   

\IEEEoverridecommandlockouts                              

\usepackage{lipsum}  
\usepackage{xcolor}
\usepackage{graphicx}
\usepackage{algorithm}
\usepackage{algpseudocode}
\usepackage{lineno,hyperref}
\usepackage{mathtools}
\usepackage{subcaption}
\usepackage{amsfonts}
\usepackage{arydshln}
\usepackage[utf8]{inputenc}
\usepackage{glossaries}
\usepackage{balance}
\usepackage{multirow}

\title{Robust Uncertainty Estimation for Classification of Maritime Objects}

\author{Jonathan Becktor, Frederik Schöller, Evangelos Boukas, and Lazaros Nalpantidis
\thanks{All authors are with the Department of Electrical and Photonics Engineering, DTU - Technical University of Denmark {\tt\small {jbibe, fets, evanb, lanalpa@dtu.dk}}}%
\thanks{This research is sponsored by the Danish Innovation Fund, The Danish Maritime Fund, Orients Fund, and the Lauritzen Foundation through the Autonomy part of the ShippingLab project, Grant number 8090-00063B.}
\thanks{Part of this research was carried out at the Jet Propulsion Laboratory, California Institute of Technology, under a contract with the National Aeronautics and Space Administration. \copyright 2023 All rights reserved.}
}

\begin{document}

\maketitle
\thispagestyle{empty}
\pagestyle{empty}

\begin{abstract}
We explore the use of uncertainty estimation in the maritime domain, showing the efficacy on toy datasets (CIFAR10) and proving it on an in-house dataset, SHIPS. We present a method joining the intra-class uncertainty achieved using Monte Carlo Dropout, with recent discoveries in the field of outlier detection, to gain more holistic uncertainty measures. We explore the relationship between the introduced uncertainty measures and examine how well they work on CIFAR10 and in a real-life setting. Our work improves the FPR95 by 8\% compared to the current highest-performing work when the models are trained without out-of-distribution data. We increase the performance by 77\% compared to a vanilla implementation of the Wide ResNet. We release the SHIPS dataset and show the effectiveness of our method by improving the FPR95 by 44.2\% with respect to the baseline. Our approach is model agnostic, easy to implement, and often does not require model retraining.
\end{abstract}


\section{Introduction}
The autonomous operation of robots, including autonomous ships, heavily relies on the perception of the world around them. Camera-based perception has seen a lot of growth in the past years. This is especially present in many well-known datasets, such as CIFAR10 and ImageNet, where models achieve almost perfect performance. However, this level of performance is often only achieved after careful reparameterization. The resulting models produce very accurate results on the specific datasets, but their performance degrades dramatically when applied to real-world robot operations, such as noisy and previously-unseen targets. Furthermore, the output of these models does not provide a usable measure of the certainty of their predictions, which is often required for robust autonomous operation. Even though the softmax function allows mapping of the logits into a probability distribution, this mapping is commonly not well calibrated, leading to over-confident predictions. While the softmax function also provides an intra-class measure of uncertainty, we will, in this work, present a method to classify samples and produce a more holistic uncertainty measure. We showcase this on a simple, commonly used dataset (CIFAR10) and on our own curated dataset (SHIPS) while also showing the performance against six outlier datasets.

\begin{figure}
    \centering
    \includegraphics[trim={0 3cm 0 3cm},clip,width=\columnwidth]{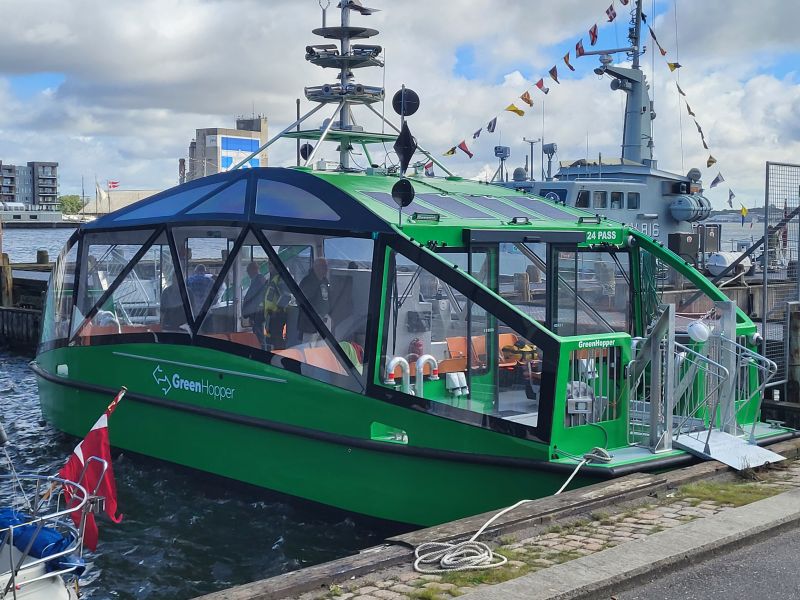}
    \caption{The GreenHopper aims to autonomously traverse "Limfjorden" in Aalborg, Denmark. Photo: Facebook Port of Aalborg, GreenHopper inauguration (Nov. 2022).}
    \label{fig:gh}
\end{figure}

Our primary focus is on autonomous operation at sea and, more precisely, on the predictive uncertainty for the classification of common maritime vessels and objects for the GreenHopper vessel, see Figure~\ref {fig:gh}. In our previous work~\cite{Becktor2020-1, Becktor2022-2, Becktor2022-3}, we proposed an object detection network tasked with robust detection of two coarse classes, \textit{buoys} and \textit{ships}; given an image, a detection consisted of an object bounding box and class confidence. This work extends our efforts of creating a more reliable and robust object detection system~\cite{Becktor2021-4, Becktor2022-5}, by focusing on producing higher quality classification outputs, that is a more precise label (e.g. from boat to sail-boat or motorboat) and providing a usable uncertainty metric for said classification.

\subsection{Contributions}
The main contributions of this paper are:
(i)~we propose a joint method to produce intra-class (aleatoric) and out-of-distribution (epistemic) uncertainty of our predictions. (ii)~We show that our method, to the best of our knowledge, performs better than any other current work at outlier detection when only trained on ID data. (iii)~We explore the relationship between the explored uncertainty measures and how they can be combined for better out-of-distribution detection.  Finally, (iv)~our work produces well-calibrated networks with usable uncertainty estimates. 

\section{Related Work}
Probabilistic inference has heavily inspired the work of network uncertainty estimates. Early work in this field, such as Bishop et al.~\cite{bishop1994mixture}, proposed using Mixture Density Networks (MDN) to estimate predictive distributions. The MDN models approximate the conditional distribution over a scalar response as a mixture of Gaussians. The parameters of a Gaussian mixture describing the predictive distribution are estimated by training a model to output parameters maximizing the overall log-likelihood. The work of~\cite{guo2017calibration} proposed calibrating the output of neural networks by scaling the logits by a constant factor before the softmax function. It showed how a calibrated network could give a better probabilistic estimate of the likelihood of a prediction. Charpentier et al.~\cite{charpentier2020posterior} estimated the latent distribution of classes to detect out-of-distribution (OOD) examples. Training a model to output the parameters of a Dirichlet distribution, it was possible to estimate predictive uncertainty with a single forward pass and classification of OOD examples. The uncertainty estimate was used to improve segmentation results on brain scans. Hendrycks et al.~\cite{hendrycks2016baseline} proposed to use the maximum softmax probability as a metric for outlier detection. The same group extended that work in~\cite{hendrycks2018deep} where they instead proposed using a subset of OOD datasets for the basis of OOD detection. Several methods for OOD detection have since been explored, such as energy score~\cite{liu2020energy}, where the authors propose to use the energy of the output vector for OOD detection. Du et al.~\cite{du2022vos} extend this by introducing a scheme that uses a multivariate distribution of a latent layer to create ``virtual outliers", which are then used for training. The work of Lee et al. introduced the Mahalanobis distance~\cite{lee2018simple} as a measure for OOD.
A method for detecting OOD examples in neural networks, dubbed ODIN, was introduced by Liang et al. in~\cite{liang2017enhancing}
which applies small perturbations to the input calibrated from an OOD sample. Hsu et al. expanded ODIN by introducing the generalized-ODIN~\cite{Hsu_2020_CVPR}, which proposes to decompose confidence scoring, removing the need to calibrate on OOD-Data.
Further work, such as the generation and collection of outlier data to be used as regularization  data for OOD detection, has been explored in several works, including \cite{lee2017training,hendrycks2018deep,mohseni2020self}. In contrast, Grcic et al. proposed to train a generative model to synthesize outliers in the pixel space~\cite{grcic2020dense}.

Producing usable uncertainty estimates for object classification is difficult; neural networks can produce a probability estimate by adding the softmax function. However, Guo et al.~\cite{pmlr-v70-guo17a} highlights the importance of model calibration, where the miss-match in class confidence and the true positive rate is often skewed towards overconfidence. Kuleshov et al.~\cite{kuleshov2018accurate} explores the use of model calibration to produce usable softmax probability estimates from networks. Lakshminarayanan et al. argue in~\cite{lakshminarayanan2017simple} that using an ensemble of models can produce well-calibrated uncertainty estimates. The ensembles are generated by training multiple instances of a model on a random permutation of a given data set. The ensemble is treated as a uniformly-weighted mixture model, and the predictions are combined using the average of the outputs. Bayesian Neural Networks (BNN)~\cite{lampinen2001bayesian}, on the other hand, define all model parameters as a Gaussian distribution, with $\mu$ and $\Sigma$ as the mean and covariance.  When training a BNN, the model parameters are updated using Bayes theorem. In practical terms, this is done by minimizing the Kullback–Leibler divergence~\cite{kullback1997information}, where a predictive uncertainty is found by sampling the model weights.

\section{Background}
The following section will outline the theory and methods used for our proposed work.
\subsection{Epistemic and Aleatoric Uncertainty}
In our work, we use the concepts of epistemic and aleatoric uncertainty \cite{smith2018understanding} as two orthogonal uncertainty estimation metrics, inspired by Wang et al.~\cite{wang2019aleatoric}.

\begin{figure*}[htb]
\centering
\includegraphics[width=1\textwidth]{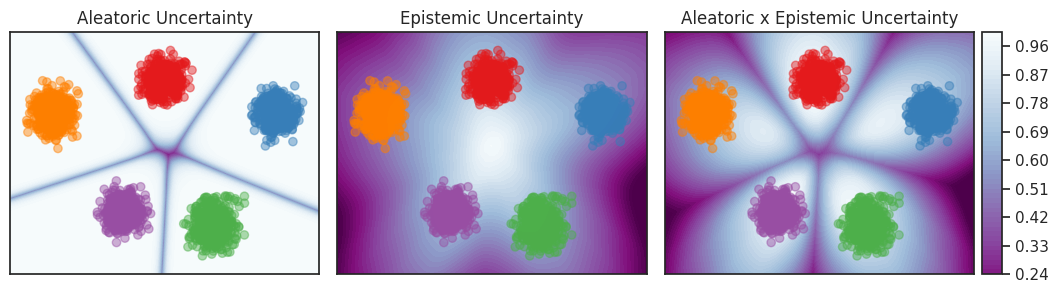}
\caption{The three plots present the probability map of the  $(x,y)$ feature space for 5 clusters. The first plot shows the intra-class (aleatoric) uncertainty, as seen in the center and between the clusters. The second plot provides the OOD (epistemic) uncertainty. The final plot shows the combination of the two providing a better uncertainty measure.}
\label{fig:rel_diagram}
\end{figure*}
Epistemic uncertainty refers to the uncertainty arising from a lack of knowledge. In machine learning, this occurs when our parameters provide an inadequate fit, often due to a lack of data, causing our posterior over parameters to be broad. Fig.~\ref{fig:rel_diagram} displays an example classification problem, where the middle plot captures epistemic uncertainty.

Aleatoric uncertainty, on the other hand, captures the stochasticity or variability in the data. Given a large dataset of high variance labels, the best possible prediction may be a high entropy one resulting in poor intra-class uncertainty. The example classification problem depicted in Fig.~\ref{fig:rel_diagram} shows the uncertainty between classes.

A more practical measure of uncertainty is the combination of epistemic and aleatoric uncertainty, as seen in the final plot of Fig.~\ref{fig:rel_diagram}. In this case, the regions that the network has not encountered are identified as uncertain regions.

\subsection{Monte Carlo Dropout}
Dropout~\cite{srivastava2014dropout} was introduced to prevent overfitting in neural networks by disabling a percentage of randomly selected neurons during training. Each neuron has some probability of being disabled, called the dropout rate. 

Monte Carlo Dropout (MC Dropout), proposed by~\cite{gal2016dropout}, allows for the use of dropout as a Bayesian approximation of the Gaussian process. The network weights are alternately dropped and simulate Monte Carlo (MC) samples from the space of all available models. 
Each dropout configuration $\Theta_{t}$ corresponds to a different sample from the approximate parametric posterior distribution $Q(\Theta|\mathcal{D})$. Thus, sampling from the approximate posterior distribution $P(T|D)$ allows for MC sampling of the model likelihood.


\subsection{Virtual Outlier Detection}
Outlier estimation estimates whether a sample is in-distribution (ID) or out-of-distribution (OOD). Our approach for estimating ID and OOD samples is based on the virtual estimation synthesis (VOS) method presented in~\cite{du2022vos}.
During training, a class conditional multivariate Gaussian for classes $c\in {1,...,C}$ is created $P_{\theta}(f(\mathbf{x}) \mid y=c)=\mathcal{N}(\boldsymbol{\mu}_{c}, \mathbf{\Sigma}_{c})$, where $\mu_{c}$ is the class mean and $\Sigma$ is the covariance. By sampling the $\epsilon-likelihood$ on the class-conditional distribution, we sample outliers $\mathcal{V}_k$.
These outliers are the basis for the work done in~\cite{du2022vos}. The outliers and the correct samples are minimized as a Binary classification problem. The goal is to reduce the energy for OOD samples and increase it for ID samples. The energy term is calculated as follows: 

\begin{equation}
\begin{aligned}
E(\mathbf{x} ; \theta)=-\log \sum_{k=1}^{K} \exp ^{f_{k}(\mathbf{x} ; \theta)}
\end{aligned}
\end{equation}

We scale the Energy term during training: $ES(\mathbf{x} ; \theta)=log(\frac{\mu_{c, ID}}{E(\mathbf{x} ; \theta)})$ to directly minimize the Energy with binary cross-entropy. To better control the VOS loss during training we compute the running mean for each class $\mu_{c, ID}$, given the respective energy $E(x;\theta)$, resulting in better regulation of class energy inconsistencies. We use the binary cross-entropy loss (BCE) on the two meta-classes, positive samples (ID) and the negative samples (OOD), drawn from the created class-conditional multivariate distributions $\hat{x}$. 

\begin{equation}
\begin{aligned}
\min _{\theta} \mathbb{E}_{(\mathbf{x}, y) \sim \mathcal{D}}[\mathcal{L}_{\text{cls}}+\beta \cdot \mathcal{L}_{\text {uncert}}]
\end{aligned}
\end{equation}
where $\mathcal{L}_{\text{cls}}$ is the Cross-Entropy Loss and $\mathcal{L}_{\text{uncert}}$ is the BCE loss, which is scaled by a factor $\beta$.

\subsection{Logit Normalization}
Logit Normalization (LN) is introduced by Wei et.al~\cite{wei2022mitigating} as a simple but effective method of reducing the number of outliers detected. This occurs by normalizing the predicted logits before computing the cross-validation loss. The loss is computed for a model $f$ given as input a target pair $(x,y)$ with weights $\theta$ as:
\begin{equation}
\begin{aligned}
    \mathcal{L}_{ln}(f(\boldsymbol{x} ; \theta), y)=-\log \frac{e^{f_y /(\tau\|\boldsymbol{f}\|)}}{\sum_{i=1}^k e^{f_i/(\tau\|\boldsymbol{f}\|)}}
\end{aligned}
\label{eq:lbl_norm}
\end{equation}
where $\tau$ is a temperature parameter that modulates the magnitude of the logits. The authors presented an impressive performance increase on an OOD dataset test suite commonly used to compare outlier detection works. The results are presented in Table~\ref{tab:method_comparison}. 

\section{Methodology}
Our goal is to provide a well-calibrated model that produces epistemic and aleatoric uncertainty estimates. We propose to sample the VOS models with MC-Dropout; this would allow for the aleatoric uncertainty to be described by MC Dropout while the Epistemic uncertainty(the OOD samples)  will be scored by the VOS Energy Score. Furthermore, with MC-Dropout, we should be able to handle extreme outliers better, as a set of Energy scores with a significant variance implies that the model is uncertain whether it is OOD or ID. We propose using the MI of the samples from MC-Dropout as a supportive OOD detection measure, as the MI of the OOD datasets appears to be lower.

To further manage the Energy Score, we introduce the Logit Normalization loss see Eq.~\ref{eq:lbl_norm}. 
By removing the incentive to increase the logits when computing the classification loss but allowing it when computing the loss on the Scaled energy, we aim to produce a network that better ensures that the magnitude of the energy is correlated to the state of a sample, whether it is ID or OOD. Furthermore, the introduction of Logit Normalization ensures calibrated networks. 

\subsection{Uncertainty measures}

Following established practices in the literature, we use the following metrics to measure the performance of detections:
\subsubsection{Aleatoric uncertainty similarity measures}
Following recent work, we here present tools for quantifying the aleatoric uncertainty.
The mutual information (MI) explored by Smith et al.~\cite{smith2018understanding} in a machine learning setting compares the predictive entropy against the expected entropy.
Entropy is calculated as follows,
\begin{equation}
\mathrm{H}(p)=-\sum_{k=1}^K p_k \log\left(p_k\right)
\end{equation}
where $p$ is the softmaxed network output, followed by the Expected $\mathbb{E}(p)=\hat{p}$. This allows us to measure the similarity between two random variables; if samples are very similar, we will get a high MI which is shown below in Eq.~\ref{eq:MI}
\begin{equation}
\operatorname{MI}=\mathrm{H}(\hat{p})-\mathbb{E}[\sum_{i=1}^{K} \mathrm{H}(p_{i})]
\label{eq:MI}
\end{equation}
where K is the number of MC samples.
It is also possible to use the Expected Kullback-Leibler Divergence (EKL):
\begin{equation}
\mathbb{E}[K L(\hat{p} \| p)]=\mathbb{E}[\sum_{i=1}^{K} \hat{p}_{i} \log (\frac{\hat{p}_{i}}{p_{i}})]
\end{equation}
This term is similar to the MI but measures the expected divergence among the possible softmax outputs.

Predictive variance is a more ad-hoc measure of uncertainty that evaluates the variance on the MC-sampled softmax outputs,
\begin{equation}
\sigma(p)=\mathbb{E}\left[(p-\hat{p})^2\right]
\end{equation}

\subsubsection{Epistemic metrics}
\begin{figure}
    \centering
    \includegraphics[width=0.85\columnwidth]{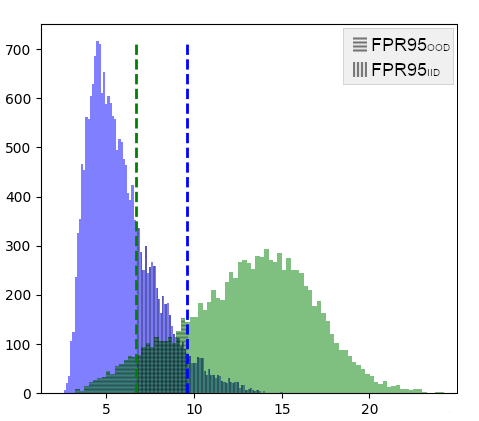}
    \caption{Comparison of OOD and ID sample histograms with the green line at 5\% quantile of ID data and the blue at 95\% quantile of OOD data. With the y-axis shows the number of samples in that bin, and the x-axis is the value of the energy score.}
    \label{fig:ood_iid_comparison}
\end{figure}

The Area Under the Receiver Operating Characteristic curve (AUROC) depicts the relationship between the True Positive Rate (TPR) and the False Positive Rate (FPR). It can be interpreted as the probability that a positive sample is assigned a higher detection score than a negative example. The AUROC score is not affected by class imbalance, which is desirable.

The Area Under the Precision-Recall curve (AUPRC), on the other hand, has a common axis, the True Positive Rate (also known as Recall) but instead maps the relationship between that and the Precision (accuracy of the model). We show this metric with respect to both the ID and OOD datasets as the positive class ($\text{AUPRC}_{ID/OOD}$).

Finally, to waive any confusion in previous works about the use of $\text{FPR95}$, we clearly define $\text{FPR95}_{ID}$ and $\text{FPR95}_{OOD}$. 
On the one hand, $\text{FPR95}_{ID}$ is used when the positive samples are the ID dataset, i.e., the number of false positive cases when $95\%$ of our data is correctly classified. 
On the other hand, the $\text{FPR95}_{OOD}$ is used when the out-of-distribution dataset is set as the positive class. When $95\%$ of positive samples are correctly classified, how many ID samples are within that range, this is visualized in figure~\ref{fig:ood_iid_comparison}.

\subsection{FPR95 discussion}
The FPR-N metric has been utilized in numerous contemporary studies on outlier detection. In certain instances, positive detection of outlier samples is demonstrated as $\text{FPR95}_{OOD}$ \cite{hendrycks2018deep, Hendrycks2019, Li_2020_CVPR, thulasidasan2021effective}, which formally introduces FPR95 as an out-of-distribution measure. However, other studies, such as those by Liu et al. \cite{liu2018open, liu2020energy, du2022vos, wei2022mitigating}, employ the in-distribution class as the positive category ($\text{FPR95}_{ID}$). Finally, Liang et al. \cite{liang2017enhancing} and Hsu et al. \cite{Hsu_2020_CVPR} use True Negative Rate at 95\% True Positive Rate (TNR@TPR95). These studies refer to one another but interchange the two methodologies for computing FPR without explicit clarification.

For $\text{FPR95}_{OOD}$, what is being tested is the proportion of ID samples with lower confidence when 95\% of the outliers are found. This is, in essence, how good we are at detecting false alarms. The $\text{FPR95}_{ID}$ metric shows us how big a proportion of the outliers are present when 95\% of the ID data is correctly classified, resulting in a metric that describes how well we find ID samples. This choice is often vague, and while the metrics are similar, it highlights different things. That is how well we predict our in-distribution samples and how good we are at finding outliers. This also applies to the AUPR metric, where we have a big difference in what is shown depending on the positive selection.

\section{Experimental Setup}
\subsection{Backbone Network Structure}
Our backbone model follows the convention set by \cite{hendrycks2018deep} that is, the Wide ResNet (WRN)~\cite{WRN2016}; this architecture strikes a balance between performance and sensitivity. The Networks are trained for 100 epochs with a cosine annealing learning rate scheduler starting at $0.1$ to produce models comparable to the VOS baseline. The Baseline WRN model achieves a 94.5\% accuracy on CIFAR10 and 97.58\% on our ships dataset. We follow the selected hyper-parameters to compare our method to the model trained in~\cite{wei2022mitigating}. Thus the models with Logit Normalization are trained for 200 epochs with an initial learning rate of 0.1 with a step-wise scheduler reducing the learning rate by a factor of ten at 80 and 140 epochs. 
Both training setups have a batch size of 128 and are optimized using SGD with a momentum of $0.9$ and weight decay of $5e-4$. When MC sampling for inference, we use a dropout with a 10\% chance of dropping a neuron.

\subsection{Datasets}
In this section, we describe the datasets used for our experiments, split into an 80/20 split of training/testing samples.

\subsubsection{CIFAR10}
For most of our experiments, we use the CIFAR10 dataset~\cite{krizhevsky2009learning} as our ID dataset. The CIFAR10 dataset consists of 60,000 32x32 color images of 10 classes: airplane, automobile, bird, cat, deer, dog, frog, horse, ship, and truck. We use this simple but popular dataset as a baseline to compare our work against previous works.

\subsubsection{OOD Data}
To test our models against outliers, we need samples that are OOD. The authors of ``ODIN"~\cite{liang2017enhancing} proposed using a set of 6 outlier datasets. We will also follow the same practice commonly followed in the related literature. Therefore, as in \cite{liang2017enhancing}, our six test datasets are: The Textures dataset, introduced by Cimpoi et al. \cite{cimpoi2014describing}, contains describable textural images. The SVHN dataset, proposed by Netzer et al. \cite{netzer2011reading}, comprises 32x32 color images of house numbers, with ten classes representing the digits 0-9. Zhou et al. \cite{zhou2017places} introduced the Places365 dataset, which comprises images for scene recognition rather than object recognition. The LSUN dataset, another scene understanding dataset, has fewer classes than Places365, and the cropped and resized versions are denoted as LSUN-C and LSUN-R, respectively. Finally, iSUN is a large-scale eye-tracking dataset selected from natural scene images in the SUN database \cite{xiao2010sun} and was introduced by Xu et al. \cite{xu2015turkergaze}.

\subsubsection{SHIPS}
This work includes a self-collected and annotated dataset. This dataset is a finely classified version of our in-house datasets and data from other sources. Stets et al. in~\cite{Stets_2019} introduced our in-house maritime dataset, which consists of 51,000 images with 31,900 annotations separated into two classes; boats and buoys. Of these, a subset has been selected and more granularly labeled. 

We include data from other relevant datasets, such as the Singapore Maritime Dataset~\cite{Prasad2017}. This dataset contains mixed samples of buoys and boats collected in Singaporean waters. Furthermore, we add data from the target area collected from online sources, primarily from videos and photos from Limfjorden (our area of interest). We add this to relevant data collected from publicly available data sets such as COCO, VOC, CIFAR10, and ImageNet.

We refer to this dataset as the \textbf{SHIPS} dataset and is made publicly available\footnote{\url{https://github.com/DTU-PAS/Ships-Classification}}. The SHIPS dataset is a classification dataset consisting of seven unique classes. Refer to Table~\ref{tab:data} for a description and examples of images of the dataset.

\begin{table}
    \begin{minipage}{0.40\linewidth}
		\centering
		\includegraphics[width=0.93\columnwidth]{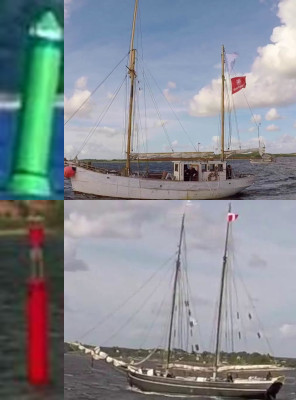}
    \label{fig:data}
	\end{minipage}
	\begin{minipage}{0.5\linewidth}
		\caption{Dataset description and example images of the SHIPS dataset}
		\label{tab:data}
		\centering
            \begin{tabular}{l|c}
            \hline\hline
            \textbf{Annotation}          & \textbf{Count T / V}\\ \hline\hline
            Red buoy                & 789 / 194         \\ 
            Green buoy              & 566 / 113         \\ 
            Human / Kayak           & 771 / 170         \\ 
            Commercial vessel       & 3.020 / 947       \\ 
            Leisure craft           & 1.915 / 476       \\ 
            Sailboat (sail down)    & 1.539 / 358       \\ 
            Sailboat (sail up)      & 2.389 / 778       \\ \hline\hline
            \end{tabular}
	\end{minipage}\hfill
\end{table}

\begin{figure}[h]
\centering
\begin{subfigure}[b]{0.48\columnwidth}
   \includegraphics[width=1\linewidth]{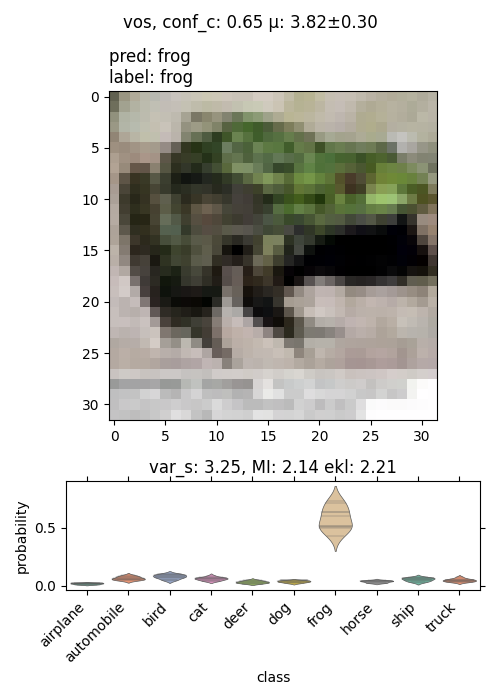}
\end{subfigure}
\begin{subfigure}[b]{0.48\columnwidth}
   \includegraphics[width=1\linewidth]{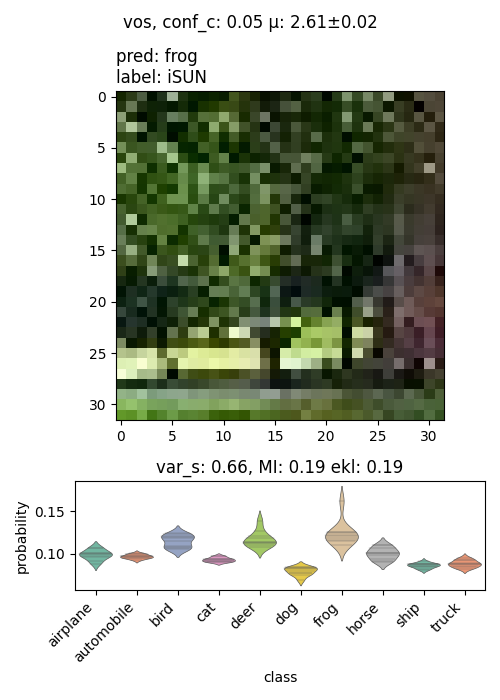}
\end{subfigure}
\caption{On the left, we have an ID sample with a high MI but low energy. On the right is an OOD sample with Low MI and low Energy.}
\label{fig:show-res}
\end{figure}

\begin{table*}[ht]
\centering
\caption{Detailed performance of the baseline vs. our method}
\label{tab:vos_detail}
\begin{tabular*}{0.9\textwidth}{l | @{\extracolsep{\fill}} ccccc}
\hline
\hline
$\text{ID}_{CIFAR10}$ & \multicolumn{5}{c|}{baseline VOS / $\text{MC}_{10}$-LN-VOS}        \\ \hline
$\text{OOD} $ & $\text{FPR95}_{ID}$ & $\text{AUPR}_{ID}$     & \multicolumn{1}{|c|}{AUROC}  & $\text{FPR95}_{OOD}$   & $\text{AUPR}_{OOD}$         \\ \hline
$Texture$       
& $46.32\pm0.80$ / $ \mathbf{24.10\pm0.76}$   & 95.93 / \textbf{98.44}    & \multicolumn{1}{|c|}{86.60 / \textbf{94.12}}
& $66.91\pm1.52$ / $ \mathbf{31.52\pm2.06}$   & 64.99 / \textbf{82.25} \\
$Places365$
& $38.18\pm1.19$ / $ \mathbf{32.39\pm0.59}$   & 97.12 / \textbf{98.45}    & \multicolumn{1}{|c|}{96.80 / \textbf{99.91}}
& $51.95\pm3.70$ / $ \mathbf{25.33\pm0.86}$   & 71.55 / \textbf{75.99}  \\ 
$LSUN_C$
& $3.01\pm0.34$  / $ \mathbf{0.13\pm0.07}$    & 99.85 / \textbf{99.57}    & \multicolumn{1}{|c|}{95.93 / \textbf{98.58}}
& $3.40\pm0.24$  / $ \mathbf{0.20\pm0.02}$    & 96.91 / \textbf{99.98}  \\
$SVHN$
& $24.85\pm0.47$ / $ \mathbf{7.20\pm0.45}$    & 99.00 / \textbf{99.70}    & \multicolumn{1}{|c|}{95.93 / \textbf{98.58}}   
& $19.23\pm0.58$ / $ \mathbf{6.90\pm0.40}$    & 80.73 / \textbf{93.88}  \\
$iSUN$ 
& $17.23\pm0.78$ / $ \mathbf{3.30\pm0.40}$    & 99.29 / \textbf{99.78}    & \multicolumn{1}{|c|}{96.52 / \textbf{98.94}} 
& $13.42\pm0.72$ / $ \mathbf{3.89\pm0.17}$    & 85.86 / \textbf{93.34}  \\
$LSUN_R$
& $14.12\pm0.73$ / $ \mathbf{3.66\pm0.43}$    & 99.42 / \textbf{99.78}    & \multicolumn{1}{|c|}{97.12 / \textbf{98.90}}
& $11.32\pm0.47$ / $ \mathbf{4.13\pm0.23}$    & 88.16 / \textbf{93.38}  \\
\hline
$Average$
& 23.96 / \textbf{11.91}            & 98.36 / \textbf{99.37}    &  \multicolumn{1}{|c|}{94.17 /\textbf{ 97.39}}
& 27.71 / \textbf{12.00}            & 81.15 / \textbf{89.97}  \\ 
\hline \hline
\end{tabular*}
\end{table*}

\begin{table*}[ht]
\centering
\caption{Comparison of models on CIFAR10 and SHIPS}
\label{tab:method_comparison}
\begin{tabular*}{0.9\textwidth}{l | @{\extracolsep{\fill}} cccccccc}
\hline \hline
$D_{ID} $   & \multicolumn{4}{c|}{CIFAR10}              & \multicolumn{4}{c}{SHIPS}     \\ \hline
$model$ & ID acc $\uparrow$     & FPR95 $\downarrow$ & AUROC $\uparrow$     & \multicolumn{1}{c|}{AUPR $\uparrow$} & ID acc $\uparrow$ & FPR95 $\downarrow$   & AUROC $\uparrow$   & AUPR $\uparrow$  \\ \hline
WRN                     
& 94.84  & 51.70 / 30.84 &  91.09     & \multicolumn{1}{c|}{97.96 / 64.57}     
& 97.58  & 54.66 / 48.62  & 87.64 & 96.32 / 54.83  \\
Baseline-VOS                     
& 94.62  & 23.96 / 27.71 &  93.86     & \multicolumn{1}{c|}{96.37 / 80.62}       
& 97.58  & 32.26 / 44.17  & 90.45     & 95.29 / 74.80  \\
$\text{MC}_{5}$-VOS    
& 94.46  & 23.53 / 26.32  & 94.31     & \multicolumn{1}{c|}{98.42 / 81.68}     
& 97.38  & 31.34 / 41.21  & 92.41 & 96.32 / 77.78  \\
$\text{MC}_{10}$-VOS    
& 94.46  & 22.91 / 25.91  & 95.09             & \multicolumn{1}{c|}{98.13 / 83.32}     
& 97.32  & \textbf{30.64 / 39.41 }& \textbf{92.41} & \textbf{96.32 / 77.98}  \\\hdashline
LN                      
& 94.50  & 14.82 / 12.90  &  96.89    & \multicolumn{1}{c|}{99.31 / 86.43}     
& 96.88  & 87.41 / 66.40 & 75.98 & 82.55 / 60.50  \\
LN-VOS                  
& 94.60  & 12.54 / 12.13 &  97.32    & \multicolumn{1}{c|}{99.36 / 89.54}     
& 96.68  & 88.99 / 68.90   & 74.12 & 80.65 / 59.50  \\
$\text{MC}_{10}$-LN-VOS  
& 94.65  & \textbf{11.91} / \textbf{12.00} & \textbf{97.39}    & \multicolumn{1}{c|}{\textbf{99.37 / 89.97}}     
& 96.68  & 76.27 / 62.50   & 79.91 &  84.45 / 69.52 \\ 
\hline \hline
\end{tabular*} 
\end{table*}

\begin{table}[ht]
\centering
\caption{Scale difference of MI for the correctly classified ID samples versus the incorrectly classified ones $\frac{F}{T}$(I). The whole ID dataset vs. the OOD datasets. We notice that the introduction of LN dramatically reduces the MI for OOD samples when training on CIFAR10, where it has the opposite effect for our SHIPS dataset $\frac{OOD}{ID}$(II).}
\label{tab:MI_comp}
\begin{tabular}{l | cc|cc}
\hline \hline
$D_{ID}$       & \multicolumn{2}{c|}{CIFAR10}              & \multicolumn{2}{c}{SHIPS}    \\ \hline
$model $          & (I)     & (II) &  (I)     & (II)   \\ \hline
$\text{MC}_{5}$-VOS         & 11.80  & 4.27  & 7.23  & 4.44   \\
$\text{MC}_{10}$-VOS        & 11.52  & 3.86  & 7.32  & 4.54   \\
\hdashline
$\text{MC}_{5}$-LN-VOS      & 1.17  & 0.38 & 1.79 & 4.32 \\ 
$\text{MC}_{10}$-LN-VOS     & 1.19  & 0.38 & 1.95 & 4.49  \\ 
\hline \hline
\end{tabular} 
\end{table}

\section{Results}
This section shows the results gathered from testing the proposed method.
Our initial results will primarily focus on our models trained on CIFAR10 to be more comparable with previous work. However, our main goal is to provide the method that has the best performance on our SHIPS dataset, which is not as curated as CIFAR10.

In Table~\ref{tab:method_comparison}, we present the improvement that VOS and MC-Dropout yield on OOD detection. We note that the MC-Sampling allows us to regulate better incorrect outlier detections that VOS provides. As we increase the number of samples for MC-Dropout, we decrease the overall FPR95; this is shown to work with diminishing returns (we found the best performance to speed was 10 MC-Samples). Furthermore, the output, when predicting with MC-Samples provides a prediction bound displaying the variability of the estimates see Figure~\ref{fig:show-res}. This is also noticeable in the energy term, where samples with significant energy variations are samples with very low confidence. Ideally, we want our samples with high confidence to have a high energy score while also having a low mutual information score. From table~\ref{tab:MI_comp}, we show that the MI for incorrectly classified ID samples is higher than the correct counterpart, per our expectations. We use the pre-trained model presented in~\cite{du2022vos} as our baseline. However, we instead sample their model with MC-Dropout by enforcing dropout at inference. This shows a slight improvement with five samples; while sampling ten times, we achieve a score of 5.0\% / 6.4\% better than the baseline VOS. Our results indicate that by using MC-sampling on a VOS model, we allow for better filtering of the incorrect OOD samples, thus achieving better $\text{FPR95}_{ID/OOD}$ and AUROC/AUPR. We see a less impressive improvement for our SHIPS dataset, with a respective improvement of 41.0\% / 10.1\% for ID and OOD going from the baseline to VOS, and with MC-Dropout, we get a 44.2\% / 20.1\% improvement.

After introducing the Log normalization loss~\cite{wei2022mitigating}, we observed a significant decrease in the $\text{FPR95}_{ID/OOD}$ compared to the baseline VOS, from $24.87$ to $11.91$ (50.3\%) and from $27.71$ to $12.00$ (57.0\%). The performance of MC-Dropout improves with increasing sample size, but with diminishing returns. However, we noticed a different outcome when applying the same testing scheme to our SHIPS dataset. The LN substantially reduced the effectiveness of OOD detection while still maintaining good accuracy and being well-calibrated. We will delve deeper into this phenomenon in the discussion.

In table~\ref{tab:MI_comp} we show the relative difference in MI between the correctly classified and incorrectly classified samples and the relative difference between the IID and OOD datasets. We can see that the inclusion of Logit Normalization decreases the logit variation in the ID datasets and the OOD ID datasets by roughly 90\%, an order of magnitude lower. We see this to a lesser extent on the SHIPS dataset (73\%). The MI difference between the ID and OOD datasets for CIFAR10 is again roughly 90\%, whereas, for the SHIPS dataset, we do not see any difference.

\section{Discussion}
The presented methods show a clear improvement in both outlier detection (epistemic uncertainty) and the improvement of intra-class confidence (aleatoric uncertainty). We want to explore issues with the current work further. We introduce new parameters that must be tuned, adding to the model complexity.
MC-Sampling requires more computation as, in essence, the model is run multiple times. These issues are of interest, and a more unified method for parameter selection and alternate methods of more efficient MC-Sampling would need to be explored.

Although not our primary goal, we show that our method can achieve, to the best of our knowledge, results not yet achieved when no OOD data is available during training. The relationship between the MI of OOD and ID samples when LN is applied remains to be adequately explored. However, the application of the LN loss on the CIFAR10 dataset reduces the MI dramatically in OOD datasets see table~\ref{tab:MI_comp}. The combination of scaled MI and the energy score, improved the performance of low-energy ID samples. Furthermore, by extending our method to include ODIN, we further improve our FPR95 results; 11.91 $\rightarrow$ 11.65 For ID and 12.00 $\rightarrow$ 10.67 OOD. 

Our findings indicate a significant disparity between the performance of models trained on CIFAR10 and SHIPS datasets. We have observed that extending OOD models with MC-Dropout is a promising approach, and we aim to investigate its potential further. Our study highlights the crucial role that dataset characteristics play in determining the effectiveness of OOD detection methods. For instance, the SHIPS dataset is relatively small and skewed towards certain types of vessels, with a limited and high entropy representation of other objects, such as buoys and humans engaged in water-based activities (e.g., kayaking, rowing, swimming). Further investigation is needed to address these challenges, how to apply LN on poorly curated datasets, and to develop robust and reliable OOD detection methods that can perform well in diverse real-world scenarios.

\section{Conclusion}
This work has aimed to explore uncertainty estimation for the classification of maritime objects. We propose a method providing a holistic and usable uncertainty measure. We have presented our experimental setup, producing well-calibrated models while providing a usable aleatoric and epistemic uncertainty measure. Furthermore, we show that our method, when considering CIFAR10, to the best of our knowledge, performs 8\% better than the previous state-of-the-art models only trained on ID data and 77\% better than the vanilla WRN model. We show the varying performance of the applied methods when considering a well-curated dataset (CIFAR10) and a more applied dataset, SHIPS. Our proposed method performs well on the CIFAR10 and SHIPS datasets, performing ~5\% better than the baseline VOS, on both datasets and 55\% / 44\% better, respectively, than the vanilla WRN models. 


\newpage
\bibliographystyle{bibstyles/IEEEtran}
\bibliography{main}

\end{document}